\pdfoutput=1
\documentclass[11pt]{article}

\usepackage{emnlp2021}

\usepackage{times}
\usepackage{latexsym}
\usepackage{microtype}
\usepackage{enumerate}
\usepackage{graphicx}
\usepackage{makecell}
\usepackage{booktabs}
\usepackage{multirow}
\usepackage{xcolor}
\usepackage{enumerate}

\usepackage[T1]{fontenc}
\usepackage[utf8]{inputenc}

\usepackage{microtype}

\title{Transformers to Fight the COVID-19 Infodemic}

\author{Lasitha Uyangodage$^\ddag$, Tharindu Ranasinghe$^	\S$, Hansi Hettiarachchi$^\heartsuit$,  \\

  $^\ddag$Department of Information Systems, University of Münster, Germany\\

  $^\S$Research Group in Computational Linguistics, University of Wolverhampton, UK \\
  
    $^\heartsuit$School of Computing and Digital Technology, Birmingham City University, UK \\

  {\tt luyangod@uni-muenster.de }}

\begin{document}
\maketitle
\begin{abstract}
The massive spread of false information on social media has become a global risk especially in a global pandemic situation like COVID-19. False information detection has thus become
a surging research topic in recent months. NLP4IF-2021 shared task on fighting the COVID-19 infodemic has been organised to strengthen the research in false information detection where the participants are asked to predict seven different binary labels regarding false information in a tweet. The shared task has been organised in three languages; Arabic, Bulgarian and English. In this paper, we present our approach to tackle the task objective using transformers. Overall, our approach achieves a 0.707 mean F1 score in Arabic, 0.578 mean F1 score in Bulgarian and 0.864 mean F1 score in English ranking 4$^{th}$ place in all the languages.
\end{abstract}

\section{Introduction}
By April 2021, coronavirus(COVID-19) pandemic has affected 219 nations around the world with 136 million total cases and 2.94 million deaths. With this pandemic situation, a rapid increase in social media usage was noticed. In measures, during 2020, 490 million new users joined indicating a more than 13\% year-on-year growth \cite{Kemp2021users}. This growth is mainly resulted due to the impacts on day-to-day activities and information sharing and gathering requirements related to the pandemic. 

As a drawback of these exponential growths, the dark side of social media is further revealed during this COVID-19 infodemic \cite{mourad2020critical}. The spreading of false and harmful information resulted in panic and confusions which make the pandemic situation worse. Also, the inclusion of false information reduced the usability of a huge volume of data which is generated via social media platforms with the capability of fast propagation. To handle these issues and utilise social media data effectively, accurate identification of false information is crucial. Considering the high data generation in social media, manual approaches to filter false information require significant human efforts. Therefore an automated technique to tackle this problem will be invaluable to the community.  

Targeting the infodemic that occurred with COVID-19, NLP4IF-2021 shared task was designed to predict several properties of a tweet including harmfulness, falseness, verifiability, interest to the general public and required attention. The participants of this task were required to predict the binary aspect of the given properties for the test sets in three languages: Arabic, Bulgarian and English provided by the organisers. Our team used recently released transformer models with the text classification architecture to make the predictions and achieved the 4$^{th}$ place in all the languages while maintaining the simplicity and universality of the method. In this paper, we mainly present our approach, with more details about the architecture including an experimental study. We also provide our code to the community which will be freely available to everyone interested in working in this area using the same methodology\footnote{The GitHub repository is publicly available on \url{https://github.com/tharindudr/infominer}}.

% Our team used recently released transformer models with the text classification architecture to make the predictions. Our approach achieved the 4$^{th}$ place in all the languages while maintaining its simplicity and universality. In this paper, we mainly present our approach, with more details about the architecture including an experimental study. We also provide our code to the community which will be freely available to everyone interested in working in this area using the same methodology\footnote{The GitHub repository is publicly available on \url{https://github.com/TharinduDR/InfoMiner}}.

\section{Related Work}
Identifying false information in social media has been a major research topic in recent years. False information detection methods can be mainly categorised into two main areas; Content-based methods and Social Context-based methods \cite{10.1145/3393880}. 

Content-based methods are mainly based on the different features in the content of the tweet. For example, \citet{10.1145/1963405.1963500} find that highly credible tweets have more URLs, and the textual content length is usually longer than that of lower credibility tweets. Many studies utilize the lexical and syntactic features to detect false information. For instance, \citet{qazvinian-etal-2011-rumor} find that the part of speech (POS) is a distinguishable feature for false information detection. \citet{6729605} find that some types of sentiments are apparent features of machine learning classifiers, including positive sentiments words (e.g., love, nice, sweet), negating words (e.g., no, not, never), cognitive action words (e.g., cause, know), and inferring action words (e.g., maybe, perhaps). Then they propose a periodic time-series model to identify key linguistic differences between true tweets and fake tweets. With the word embeddings and deep learning getting popular in natural language processing, most of the fake information detection methods were based on embeddings of the content fed into a deep learning network to perform the classification \cite{10.5555/3061053.3061153}. 

Traditional content-based methods analyse the credibility of the single microblog or claim in isolation, ignoring the high correlation between different tweets and events. However, Social Context-based methods take different tweets in a user profile or an event to identify false information. Many studies detect false information by analyzing users’ credibility \cite{li-etal-2019-rumor} or stances \cite{10.1145}. Since this shared is mainly focused on the content of the tweet to detect false information, we can identify our method as a content-based false information identification approach.

\section{Data}
\label{sec:data}
The task is about predicting several binary properties of a tweet on COVID-19: whether it is harmful, whether it contains a verifiable claim, whether it may be of interest to the general public, whether it appears to contain false information, etc. 
\cite{NLP4IF-2021-COVID19-task}. The data has been released for three languages; English, Arabic and Bulgarian \footnote{The dataset can be downloaded from \url{https://gitlab.com/NLP4IF/nlp4if-2021}}. Following are the binary properties that the participants should predict for a tweet. 

\begin{enumerate}[I]
  \item \textbf{Verifiable Factual Claim}: Does the tweet contain a verifiable factual claim?
  \item \textbf{False Information}: To what extent does the tweet appear to contain false information?
  \item \textbf{Interest to General Public}: Will the tweet have an effect on or be of interest to the general public?
  \item \textbf{Harmfulness}: To what extent is the tweet harmful to the society?
  \item \textbf{Need of Verification}: Do you think that a professional fact-checker should verify the claim in the tweet?
  \item \textbf{Harmful to Society}: Is the tweet harmful for the society?
  \item \textbf{Require attention}: Do you think that this tweet should get the attention of government entities?

\end{enumerate}

\section{Architecture}
The main motivation for our architecture is the recent success that the transformer models had in various natural language processing tasks like sequence classification \cite{ranasinghe-hettiarachchi-2020-brums, ranasinghe2019brums, pitenis-etal-2020-offensive}, token classification \cite{mudes, ranasinghe2021semeval}, language detection \cite{jauhiainen2021}, word context prediction \cite{hettiarachchi-ranasinghe-2020-brums, hettiarachchi2021semeval} question answering \cite{yang-etal-2019-end-end-open} etc. Apart from providing strong results compared to RNN based architectures \cite{hettiarachchi-ranasinghe-2019-emoji, ranasinghe2019brums}, transformer models like BERT \cite{devlin-etal-2019-bert} provide pretrained multilingual language models that support more than 100 languages which will solve the multilingual issues of these tasks \cite{ranasinghe2020wlv, ranasinghe2021tallip, ranasinghe-zampieri-2020-multilingual}. 

\begin{figure}[ht]
\centering
\includegraphics[scale=0.4]{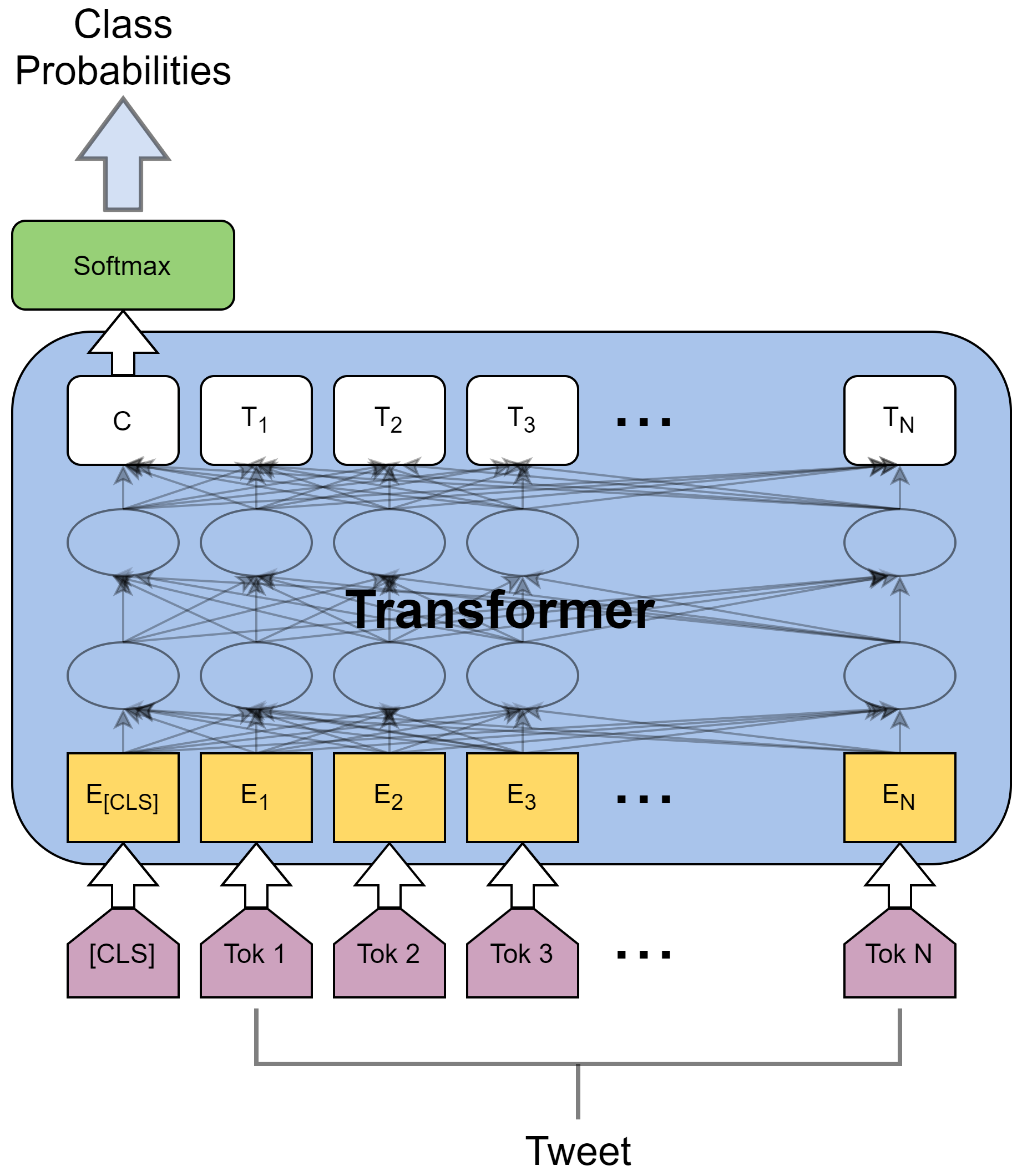}
\caption{Text Classification Architecture}
\label{fig:architecture}
\end{figure}

For sequence classification tasks transformer models take an input of a sequence and outputs the representations of the sequence. There can be one or two segments in a sequence which are separated by a special token [SEP] \cite{devlin-etal-2019-bert}. In this approach we considered a tweet as a sequence and no [SEP] token is used. Another special token [CLS] is used as the first token of the sequence which contains a special classification embedding. For text classification tasks, transformer models take the final hidden state $\textbf{h}$ of the [CLS] token as the representation of the whole sequence \cite{10.1007/978-3-030-32381-3_16}. A simple softmax classifier is added to the top of the transformer model to predict the probability of a class $c$ as shown in Equation \ref{equ:softmax} where $W$ is the task-specific parameter matrix. In the classification task all the parameters from transformer as well as W are fine tuned jointly by maximising the log-probability of the correct label. The architecture of transformer-based sequence classifier is shown in Figure \ref{fig:architecture}.

\begin{equation}
\label{equ:softmax}
p(c|\textbf{h}) = softmax(W\textbf{h}) 
\end{equation}

\renewcommand{\arraystretch}{1.2}
\begin{table*}[!ht]
\begin{center}
\small
\scalebox{0.95}{
%\footnotesize
\begin{tabular}{l l c c c c c c c | c} 
%\hline
\toprule

&{\bf \makecell{Model} } & \makecell{\bf I} & \makecell{\bf II} & \makecell{\bf III} & \makecell{\bf IV} & \makecell{\bf V} & \makecell{\bf VI} & \makecell{\bf VII} & \makecell{\bf Mean} \\
\midrule
\multirow{2}{*}{\bf English} 
& roberta-base & 0.822 & 0.393 & 0.821 & 0.681 & 0.461 & 0.235 & 0.251 & 0.523 \\

& bert-base-cased & 0.866 & 0.461 & 0.893 & 0.740 & 0.562 & 0.285 & 0.303 & 0.587 \\

\midrule
\multirow{3}{*}{\bf Arabic} 
& bert-multilingual-cased & 0.866 & 0.172 & 0.724 & 0.400 & 0.557 & 0.411 & 0.625 & 0.536 \\
& arabert-v2 & 0.917 & 0.196 & 0.782 & 0.469 & 0.601 & 0.433 & 0.686 & 0.583 \\
& arabert-v2-tokenized & 0.960 & 0.136 & 0.873 & 0.571 & 0.598 & 0.424 & 0.678 & 0.606 \\

\midrule

\multirow{1}{*}{\bf Bulgarian} & 
bert-multilingual-cased & 0.845 & 0.098 & 0.516 & 0.199 & 0.467 & 0.303 & 0.196 & 0.375  \\

\bottomrule
%\bottomrule
\end{tabular}
}
\end{center}
\caption{Macro F1 between the algorithm predictions and human annotations for development set in all the languages. Results are sorted from Mean F1 score for each language.} 
\label{tab:results}
\end{table*}

\renewcommand{\arraystretch}{1.2}
\begin{table*}[!ht]
\begin{center}
\small
\scalebox{0.95}{
%\footnotesize
\begin{tabular}{l l c c c c c c c | c} 
%\hline
\toprule

&{\bf \makecell{Model} } & \makecell{\bf I} & \makecell{\bf II} & \makecell{\bf III} & \makecell{\bf IV} & \makecell{\bf V} & \makecell{\bf VI} & \makecell{\bf VII} & \makecell{\bf Mean} \\
\midrule
\multirow{3}{*}{\bf English} 
& Best System & 0.835 & 0.913 & 0.978 & 0.873 & 0.882 & 0.908 & 0.889 & 0.897 \\
& InfoMiner & 0.819 & 0.886 & 0.946 & 0.841 & 0.803 & 0.884 & 0.867 & 0.864 \\
& Random Baseline & 0.552 & 0.480 & 0.457 & 0.473 & 0.423 & 0.563 & 0.526 & 0.496\\

\midrule
\multirow{3}{*}{\bf Arabic} 
& Best System & 0.843 & 0.762 & 0.890 & 0.799 & 0.596 & 0.912 & 0.663 & 0.781 \\
& InfoMiner & 0.852 & 0.704 & 0.774 & 0.743 & 0.593 & 0.698 & 0.588 & 0.707 \\
& Random Baseline & 0.510 & 0.444 & 0.487 & 0.442 & 0.476 & 0.584 & 0.533 & 0.496 \\

\midrule

\multirow{3}{*}{\bf Bulgarian} 
& Best System & 0.887 & 0.955 & 0.980 & 0.834 & 0.819 & 0.678 & 0.706 & 0.837 \\
& InfoMiner & 0.786 & 0.749 & 0.419 & 0.599 & 0.556 & 0.303 & 0.631 & 0.578 \\
& Random Baseline & 0.594 & 0.502 & 0.470 & 0.480 & 0.399 & 0.498 & 0.528 & 0.496 \\

\bottomrule
%\bottomrule
\end{tabular}
}
\end{center}
\caption{Macro F1 between the InfoMiner submission and human annotations for test set in all the languages. Best System is the results of the best model submitted for each language as reported by the task organisers \cite{NLP4IF-2021-COVID19-task}.} 
\label{tab:results_test}
\end{table*}

\section{Experimental Setup}
We considered the whole task as seven different classification problems. We trained a transformer model for each label mentioned in Section \ref{sec:data}. This gave us the flexibility to fine-tune the classification model in to the specific label rather than the whole task. Given the very unbalanced nature of the dataset, the transformer models tend to overfit and predict only the majority class. Therefore, for each label we took the number of instances in the training set for the minority class and undersampled the majority class to have the same number of instances as the minority class. 

We then divided this undersampled dataset into a training set and a validation set using 0.8:0.2 split. We mainly fine tuned the learning rate and number of epochs of the classification model manually to obtain the best results for the development set provided by organisers in each language. We obtained $1e^{-5}$ as the best value for learning rate and 3 as the best value for number of epochs for all the languages in all the labels. The other configurations of the transformer model were set to a constant value over all the languages in order to ensure consistency between the languages. We used a batch-size of eight, Adam optimiser \cite{kingma2014adam} and a linear learning rate warm-up over 10\% of the training data. The models were trained using only training data. We performed early stopping if the evaluation loss did not improve over ten evaluation rounds. A summary of hyperparameters and their values used to obtain the reported results are mentioned in Table \ref{tab:params}. The optimized hyperparameters are marked with $\ddag$ and their optimal values are reported. The rest of the hyperparameter values are kept as constants. We did not use any language specific preprocessing techniques in order to have a flexible solution between the languages. We used a Nvidia Tesla K80 GPU to train the models. All the experiments were run for five different random seeds and as the final result, we took the majority class predicted by these different random seeds as mention in \citet{hettiarachchi-ranasinghe-2020-infominer}. We used the following pretrained transformer models for the experiments.

\paragraph{bert-base-cased} - Introduced in \citet{devlin-etal-2019-bert}, the model has been trained on a Wikipedia dump of English using Masked Language Modelling (MLM) objective. There are two variants in English BERT, base model and the large model. Considering the fact that we built seven different models for each label, we decided to use the base model considering the resources and time. 

\paragraph{roberta-base} - Introduced in \citet{liu2019roberta}, RoBERTa builds on BERT and modifies key hyperparameters, removing the next-sentence pretraining objective and training with much larger mini-batches and learning rates. RoBERTa has outperformed BERT in many NLP tasks and it motivated us to use RoBERTa in this research too. Again we only considered the base model.

\paragraph{bert-nultilingual-cased} - Introduced in \citet{devlin-etal-2019-bert}, the model has been trained on a Wikipedia dump of 104 languages using MLM objective. This model has shown good performance in variety of languages and tasks. Therefore, we used this model in Arabic and Bulgarian.

\paragraph{AraBERT}
 Recently language-specific BERT based models have proven to be very efficient at language understanding. AraBERT \cite{antoun-etal-2020-arabert} is such a model built for Arabic with BERT using scraped Arabic news websites and two publicly available Arabic corpora; 1.5 billion words Arabic Corpus \cite{elkhair201615} and OSIAN: the Open Source International Arabic News Corpus \cite{zeroual-etal-2019-osian}. Since AraBERT has outperformed multilingual bert in many NLP tasks in Arabic \cite{antoun-etal-2020-arabert} we used this model for Arabic in this task. There are two version in AraBERT; AraBERTv0.1 and AraBERTv1, with the difference being that AraBERTv1 uses pre-segmented text where prefixes and suffixes were splitted using the Farasa Segmenter \cite{abdelali-etal-2016-farasa}.

\section{Results}

When it comes to selecting the best model for each language, highest F1 score out of the evaluated models was chosen. Due to the fact that our approach uses a single model for each label, our main goal was to achieve good F1 scores using light weight models. The limitation of available resources to train several models for all seven labels itself was a very challenging task to the team but we managed to evaluate several. 

% As we mentioned earlier, in order to avoid models getting over/under fitting to the test data, we had to under sample the data to get the classes balanced which drastically reduced the number of records for some of the labels and ended up resulting very low F1 scores. 

As depicted in Table \ref{tab:results}, for English, bert-base-cased model performed better than roberta-base model. For Arabic, arabert-v2-tokenized  performed better than the other two models we considered. For Bulgarian, with the limited time, we could only train bert-multilingual model, therefore, we submitted the predictions from that for Bulgarian. 

As shown in Table \ref{tab:results_test}, our submission is very competitive with the best system submitted in each language and well above the random baseline. Our team was ranked 4$^{th}$ in all the languages.

\section{Conclusion}
We have presented the system by InfoMiner team for NLP4IF-2021-Fighting the COVID-19 Infodemic. We have shown that multiple transformer models trained on different labels can be successfully applied to this task. Furthermore, we have shown that undersampling can be used to prevent the overfitting of the transformer models to the majority class in an unbalanced dataset like this. Overall, our approach is simple but can be considered as effective since it achieved 4$^{th}$ place in the leader-board for all three languages.

One limitation in our approach is that it requires maintaining seven transformer models for the seven binary properties of this task which can be costly in a practical scenario which also restricted us from experimenting with different transformer types due to the limited time and resources. Therefore, in future work, we are interested in remodeling the task as a multilabel classification problem, where a single transformer model can be used to predict all seven labels.

\section*{Acknowledgments}

We would like to thank the shared task organizers for making this interesting dataset available. We further thank the anonymous reviewers for their insightful feedback.

\bibliography{anthology}
\bibliographystyle{acl_natbib}

%\appendix
\appendix
\section{Appendix}

A summary of hyperparameters and their values used to obtain the reported results are mentioned in Table \ref{tab:params}. The optimised hyperparameters are marked with $\ddag$ and their optimal values are reported. The rest of the hyperparameter values are kept as constants. 

\renewcommand{\arraystretch}{1.2}
\begin{table}[!ht]
\centering
\scalebox{1.0}{
\small
\begin{tabular}{ll}
\toprule
\bf Parameter &   \bf Value \\ \hline
learning rate$^{\ddag}$ & $1e^{-5}$\\
number of epochs$^{\ddag}$ & $3$\\
adam epsilon & $1e^{-8}$\\
warmup ration & 0.1\\
warmup steps & 0\\
max grad norm & 1.0\\
max seq. length & 120\\
gradient accumulation steps & 1\\
\bottomrule
\end{tabular}
}
\caption{Hyperparameter specifications}
\label{tab:params}
\end{table}

\end{document}